\documentclass[unnumsec,webpdf,contemporary,large]{oup-authoring-template}%
\usepackage{booktabs}
\usepackage{tikz}
\usepackage{pgfplots}
\usepackage{graphicx}
\usepackage{hyperref}

\theoremstyle{thmstyleone}%
\theoremstyle{thmstyletwo}%
\theoremstyle{thmstylethree}%

\begin{document}

\journaltitle{ADRS-CNet: An adaptive dimensionality reduction selection and classification network for DNA storage clustering algorithms}
\DOI{DOI HERE}
\copyrightyear{2022}
\pubyear{2019}
\access{Advance Access Publication Date: Day Month Year}
\appnotes{Paper}

\firstpage{1}
\graphicspath{{Fig/}}

%\subtitle{Subject Section}

\title[ADRS-CNet: An adaptive dimensionality reduction selection and classification network for DNA storage clustering algorithms]{ADRS-CNet: An adaptive dimensionality reduction selection and classification network for DNA storage clustering algorithms}

\author[1]{Bowen Liu \ORCID{0009-0001-4934-327X} \thanks{These authors contributed equally to this work.}}
\author[2]{Jiankun Li \thanks{These authors contributed equally to this work.}}
%\author[3]{Vuong Phan \thanks{Corresponding author.}}
\author[4]{Xiang Peng}
\author[5]{Xiangrun Zheng}

\authormark{Bowen Liu et al.}

\address[1]{\orgdiv{School of Mathematical Sciences}, \orgname{University of Southampton}, \orgaddress{\street{University Road}, \postcode{SO17 1BJ}, \state{Southampton}, \country{United Kingdom}}}
\address[2]{\orgdiv{School of Mathematical Sciences}, \orgname{University of Southampton}, \orgaddress{\street{University Road}, \postcode{SO17 1BJ}, \state{Southampton}, \country{United Kingdom}}}
%\address[3]{\orgdiv{School of Mathematical Sciences}, \orgname{University of Southampton}, \orgaddress{\street{54/10005}, \postcode{SO17 1TR}, \state{Southampton}, \country{United Kingdom}}}
\address[4]{\orgdiv{School of Electronics and Computer Science}, \orgname{University of Southampton}, \orgaddress{\street{University Road}, \postcode{SO17 1BJ}, \state{Southampton}, \country{United Kingdom}}}
\address[5]{\orgdiv{School of Mathematical Sciences}, \orgname{University of Southampton}, \orgaddress{\street{University Road}, \postcode{SO17 1BJ}, \state{Southampton}, \country{United Kingdom}}}

%\corresp[3]{Corresponding author. \href{email:t.v.phan@soton.ac.uk}{t.v.phan@soton.ac.uk}}

\received{Date}{0}{Year}
\revised{Date}{0}{Year}
\accepted{Date}{0}{Year}

%\editor{Associate Editor: Name}

%\abstract{
%\textbf{Motivation:} .\\
%\textbf{Results:} .\\
%\textbf{Availability:} .\\
%\textbf{Contact:} \href{name@email.com}{name@email.com}\\
%\textbf{Supplementary information:} Supplementary data are available at \textit{Journal Name}
%online.}

\abstract{DNA storage technology offers new possibilities for addressing massive data storage due to its high storage density, long-term preservation, low maintenance cost, and compact size. To improve the reliability of stored information, base errors and missing storage sequences are challenges that must be faced. Currently, clustering and comparison of sequenced sequences are employed to recover the original sequence information as much as possible. Nonetheless, extracting DNA sequences of different lengths as features leads to the curse of dimensionality, which needs to be overcome. To address this, techniques like PCA, UMAP, and t-SNE are commonly employed to project high-dimensional features into low-dimensional space. Considering that these methods exhibit varying effectiveness in dimensionality reduction when dealing with different datasets, this paper proposes training a multilayer perceptron model to classify input DNA sequence features and adaptively select the most suitable dimensionality reduction method to enhance subsequent clustering results. Through testing on open-source datasets and comparing our approach with various baseline methods, experimental results demonstrate that our model exhibits superior classification performance and significantly improves clustering outcomes. This displays that our approach effectively mitigates the impact of the curse of dimensionality on clustering models.}

% \boxedtext{
% \begin{itemize}
% \item Key boxed text here.
% \item Key boxed text here.
% \item Key boxed text here.
% \end{itemize}}

\maketitle

\section{Introduction}

In recent years, the proliferation of new internet devices and the surge in demand for their services have led to the generation and collection of vast amounts of data \cite{b9}. According to the International Data Corporation (IDC), global data storage demand is projected to reach 175 ZB by 2025 \cite{b11}. However, the maximum density of current storage media is only 103 GB/mm\textsuperscript{3} \cite{b12}, which is insufficient to meet future data storage needs. Moreover, with the increasing storage demand, we will encounter significant challenges related to maintenance costs, data transmission expenses, and the lifespan of storage media \cite{b13} \cite{b14}. Consequently, it has become imperative to develop new information storage solutions to address the escalating data storage requirements. 

DNA carries the genetic information of all organisms, composed of four nucleotide bases: A, T, C, and G \cite{b19}. Its storage technology utilizes artificially synthesized DNA as a medium for information storage. Due to advantages like high storage density, long preservation time, small volume, and low maintenance cost, DNA storage technology provides a "breakthrough" possibility for solving the storage and application challenges of massive data \cite{b17} \cite{b10} \cite{b9}.

In the downstream information retrieval process of DNA storage technology, specific hybridization techniques, such as Polymerase Chain Reaction (PCR) or magnetic bead separation technology, are commonly used to access data  \cite{b3}. However, this technology faces several challenges, including high base errors (insertions, deletions, substitutions, etc. ) and missing storage sequences, which pose significant threats to the reliability of the stored data \cite{b20}. 

To solve these issues, clustering and alignment of sequenced data can be employed. A popular method is based on k-mer frequency matrices, where the dimensionality of the extracted features increases exponentially with the value of k \cite{b15} \cite{b16} \cite{b24}. Therefore, selecting an appropriate dimensionality reduction technique becomes a critical challenge that needs to be addressed. 

Among the array of algorithms available, Principal Component Analysis (PCA) \cite{b7}, t-distributed Stochastic Neighbor Embedding (t-SNE) \cite{b22}, and Uniform Manifold Approximation and Projection (UMAP) \cite{b5} are notably prominent in the fields of cell biology, bioinformatics, and data visualization \cite{b8}.This study intends to develop an adaptive classification model (ADRS-CNet) to find the optimal dimensionality reduction methods to mitigate the curse of dimensionality caused by k-mer extraction of sequencing features, thereby improving the effectiveness of K-means clustering and restoring the original sequence information as much as possible.  

%Among the array of algorithms available, Principal Component Analysis (PCA) \cite{b7}, t-distributed Stochastic Neighbor Embedding (t-SNE) \cite{b22}, and Uniform Manifold Approximation and Projection (UMAP) \cite{b5} are notably prominent in the fields of cell biology, bioinformatics, and data visualization \cite{b8}. This study tackles the challenge of selecting suitable dimensionality reduction methods to mitigate the curse of dimensionality in K-means clustering. The study involves extracting a comprehensive set of features from DNA sequence datasets using k-mer frequency matrices. Dimensionality reduction of the feature datasets is sequentially achieved by applying PCA, UMAP, and t-SNE. The effectiveness of each method is evaluated based on clustering accuracy, allowing us to identify and categorize the most appropriate method for each dataset. This categorization process, referred to as dataset sub-tagging, organizes the labeled datasets into a cohesive group. This group is subsequently used to train a multilayer perceptron (MLP) \cite{b51} model to achieve model adaptability.
\section{Related Works}

DNA storage technology, as a cutting-edge information storage method, can be divided into three main stages: upstream, midstream, and downstream. These stages include six critical steps \cite{b3}: encoding, synthesis, storage, retrieval, sequencing and decoding (Refer to Fig.1).\\Errors can occur during DNA sequencing, complicating the decoding process. Robust mechanisms are required to detect and correct these errors.  Two primary approaches are employed: using error correction codes alone, and combining clustering methods with error correction codes to enhance error correction \cite{b32}. 

\subsection{\textbf{Error Correction Coding} }
Shubham Chandak \cite{b34} developed a novel scheme using a single large block-length LDPC code for both error and erasure correction, introducing new techniques to handle insertion and deletion errors during the synthesis process. As a result, their approach, tested with experimental data obtained through array synthesis and Illumina sequencing, achieved a 30-40\% reduction in reading costs, significantly enhancing the efficiency and reliability of DNA storage systems. 

The team led by Yaniv Erlich \cite{b33} utilized the DNA Fountain storage strategy, incorporating an efficient error-correction code system, to perfectly retrieve information from the sequencing coverage equivalent to a single Illumina sequencing tile. They tested a process allowing for $2.18 \times 10^{15}$ retrievals with accurate data decoding. Furthermore, they explored the storage density limits, achieving $215$ PB per gram of DNA, significantly surpassing previous records.

However, using error correction codes alone to correct sequencing errors can have limitations in certain cases. In this context, if encoded data is preliminarily reconstructed from sequenced raw data using clustering techniques and then decoded with error correction codes, the probability of recovering the original sequences increases while reducing sequencing redundancy, costs, and complexity \cite{b32}. 
\subsection{\textbf{Clustered Error Correction Coding} }
Clustered Error Correction Coding integrates clustering and error correction codes, providing a comprehensive approach. For clustering, it primarily encompasses sequence clustering methods utilizing alignment techniques and clustering approaches that employ k-mer counting.

\subsubsection{\textbf{Alignment-Based Clustering Methods}}

Weizhong Li and Adam Godzik \cite{b35} developed CD-HIT, a program that significantly accelerates the clustering of protein or nucleotide sequences by using a short word filtering method to bypass full sequence alignment. This algorithm effectively clusters sequences by comparing short subsequences, enabling it to process large databases containing millions of sequences more efficiently, which enhances computational performance while maintaining accuracy.

An exact algorithm to determine which pairs of sequences lie within a given Levenshtein distance was developed by Eduard Zorita \cite{b36}, who named this algorithm ‘Starcode’. This algorithm is particularly useful for correcting sequencing errors or reducing redundancy in data. More precisely, they introduced a novel implementation called poucet search, which applies the Needleman–Wunsch algorithm \cite{b37} on the nodes of a trie. In practical applications, this algorithm excels in matching random DNA barcodes, significantly improving both speed and precision compared to existing sequence clustering algorithms. 

A classic approach is to use a greedy method \cite{b38} \cite{b35} to maximize the performance of biological sequence clustering. However, it is noteworthy that the sequence insertion order is highly sensitive to the results. Without appropriate size sorting or abundance sorting, the selection of cluster centers may not be precise, resulting in a negative effect on the overall clustering effect \cite{b40}. Although optimal sorting can theoretically solve this problem, achieving optimal ordering requires knowing the correct cluster centroids in advance \cite{b41}, which is not feasible. 

\subsubsection{\textbf{K-mer Counting-Based Clustering Methods}}
Many bioinformatics clustering algorithms rely on methods such as the Needleman-Wunsch algorithm \cite{b37}, a classic global alignment technique designed to find the optimal alignment between two sequences. The algorithm evaluates the quality of alignments through similarity metrics and employs a dynamic programming algorithm to compute the optimal solution \cite{b43}. However, its computational cost,  $O(mn)$ \cite{b43} (where m and n are the lengths of the sequences being compared), becomes impractical with massive datasets. Therefore, in modern sequence retrieval, it is necessary to integrate more efficient comparison tools to balance computational efficiency and alignment accuracy. A representative example is the alignment-free clustering method based on k-mer counting for genomic sequences \cite{b44}.

Bao et al.  \cite{b45} developed an entropy-based alignment-free model known as Category-Position-Frequency (CPF), which enhances clustering performance through a 12-dimensional feature vector. This model leverages k-mer technology and adjustable sliding window lengths, demonstrating high efficiency and versatility across various datasets.

James et al. \cite{b46} introduced MeShClust, an unsupervised learning tool for DNA sequence clustering that utilizes k-mer frequencies and the mean shift algorithm. It employs a Generalized Linear Model (GLM) to identify sequence similarities and optimizes clustering through an iterative process. Compared to existing tools, MeShClust demonstrates superior efficiency and accuracy.

Yana Hrytsenko et al. \cite{b47} conducted a study to determine population structure using k-mer frequencies and principal component analysis (PCA). They transformed DNA sequences into k-mer frequency matrices and applied PCA and K-means clustering to these matrices. The results showed that this approach effectively identified population structures and differentiated admixed from non-admixed individuals, demonstrating comparable accuracy to traditional model-based methods. 

Our research is inspired by the findings of Yana Hrytsenko et al. \cite{b47} and Yuta Hozumi et al. \cite{b8}. Through exploring widely used dimensionality reduction techniques in bioinformatics, including UMAP, t-SNE, and PCA, we have developed an adaptive model to select the optimal dimensionality reduction method depends on specific data structures. Integrating this model with k-mer frequency matrices and K-means clustering, we apply it to error analysis in DNA storage technologies. Our approach involves increasing the dimensions of k-mer frequency matrices to enhance feature capture rates and combining this with optimal dimensionality reduction techniques to mitigate the impact of sparse matrices on traditional K-means clustering models. This method aims to alleviate the curse of dimensionality in K-means frequency matrices and avoid the significant computational costs associated with distance calculations for DNA sequence alignments.

\section{Methodology}\label{sec3}
\subsection{\textbf{Feature Engineering}}

K-means clustering, as a method reliant on feature extraction, crucially depends on the identification of specific sequence patterns within DNA sequences to determine their functional characteristics, thereby influencing sequence clustering outcomes significantly \cite{b6}. However, the inherent variability in DNA sequence lengths due to insertions, deletions, substitutions, and breaks during PCR amplification poses challenges for traditional clustering approaches such as K-means clustering. To mitigate these challenges, sequences are often subjected to truncation or padding. K-mer counting stands as a traditional and widely adopted solution extensively utilized in the feature extraction process of DNA sequences \cite{b19}. 

K-mer frequency matrix works by sliding a fixed-length window of size k along DNA or amino acid sequences, enumerating all possible k-mers, and counting the frequency of each k-mer's occurrence in the sequences. This process generates a matrix where each row represents a sequence and each column represents a possible k-mer, with each element in the matrix indicating the frequency of the corresponding k-mer in that sequence. By employing this method, we can handle unaligned DNA sequences and use each column of the frequency matrix as a feature for clustering, utilizing traditional clustering methods for analysis. 
 
In this study, while opting for larger word lengths (the value of k) can enhance the accuracy and effectiveness of clustering algorithms, we must carefully consider the resultant dimensionality explosion (with a magnification factor of \({ 4 } ^{ n  }\), where n represents the word length \cite{b24}) and the costly computational overhead associated with t-SNE dimensionality reduction techniques in high-dimensional spaces. Therefore, we constructed all possible nucleotide pair permutations as features under a given DNA feature length of k=5, and traversed all DNA sequence lengths to count the frequency of appearance of each k-mer in the sequenced sequences.

\subsection{\textbf{K-means Clustering Algorithm}}

The K-means clustering algorithm is one of the most powerful and popular data mining algorithms in the research community. The algorithm achieves the minimum distance between clusters by minimizing the squared distance between each observation point and the centroid of its respective cluster. For each observation point, the distance to the centroid of each cluster is calculated and the point is assigned to the nearest cluster. 

Given a dataset \( \mathbf{X} = \{\mathbf{x}_1, \mathbf{x}_2, \ldots, \mathbf{x}_n\} \), these data points will be partitioned into \( k \) clusters \( C_1, C_2, \ldots, C_k \), such that the points within each cluster are as close as possible (minimum Euclidean distance).
\\The calculation formula of Euclidean distance is as follows:

\begin{equation}
\displaystyle\sum_{j=1}^{k} \sum\nolimits_{{x}_{i}\epsilon {C}_{j}}^{}{\begin{Vmatrix}{x}_{i}-{\mu }_{i}

\end{Vmatrix}}_{2}^{2}\label{eq1}
\end{equation}
Where, \({\begin{Vmatrix}\bullet 

\end{Vmatrix}}_{2}\) express \({l}_{2}\) standards, and \({\mu }_{i}\) is the mean of the data points in cluster \textit{j}.

In order to ensure that the centre represents the average position of the set value inside the pivot, we need to update the centre of mass by calculating the instructions for all assigned observation points for each pivot, setting it as the new pivot. The K-means algorithm iteratively updates the cluster centers to minimum the objective function (Equation 1).

\subsection{\textbf{UMAP}}

Uniform Manifold Approximation and Projection (UMAP) is a novel manifold learning technique capable of capturing the nonlinear structure in data and exhibiting superior runtime performance \cite{b4}. 

This paper explores the principles, implementation, and applications of UMAP from a computational perspective. Further theoretical details can be found in the referenced literature \cite{b5}. The process of UMAP dimensionality reduction can be roughly divided into three stages:

\begin{itemize}
\item \textbf{Constructing high-dimensional fuzzy topology} 
\\In the UMAP dimensionality reduction process, high-dimensional data points are represented by  \(X=\begin{Bmatrix}
{x}_{1},...,{x}_{n}
\end{Bmatrix}\) and low-dimensional data points are represented by  \(Y=\begin{Bmatrix}
{y}_{1},...,{y}_{n}

\end{Bmatrix}\). Let \(d:X\ast X\rightarrow {\mathbb{R}}^{+ }\) denote the metric space, where given a hyperparameter \textit{k}, the nearest neighbor set \( \begin{Bmatrix}
{x}_{i1},...,{x}_{in}
\end{Bmatrix}\)  for \({x}_{i}\) under \textit{d} can be obtained. The high-dimensional fuzzy topology can then be represented using an exponential probability distribution as follows: 
\begin{equation}
{p}_{i|j}= {e}^{-\frac{d({x}_{i},{x}_{j})-{\rho }_{j}}{{\sigma }_{i}}}
\end{equation}
 Where, \({\rho }_{j}\) and \({\sigma }_{i}\) represents adjusting parameters, which can be determined through the following equation:
 \begin{equation}
k = {2}^{\textstyle\sum_{i}^{}{p}_{ij}}\label{eq3}
\end{equation}
To ensure its symmetry, the following formula is used for transformation:
\begin{equation}
    {p}_{ij}={p}_{i|j}+ {p}_{j|i}-{p}_{i|j}{p}_{j|i}\label{eq4}
\end{equation}
\item \textbf{Constructing probability distribution in the lower-dimensional space} 
\begin{equation}
    {q}_{i}=(1+{a({y}_{i}-{y}_{j})}^{2b})^{-1}\label{eq5}
\end{equation}
Where, by adjusting parameters \(a\) and \(b\), the aggregation of the data after mapping can be controlled.
\item \textbf{Loss unction} 
\\In order to preserve the structural characteristics of the dataset as much as possible, the UMAP algorithm applies attractive forces to similar data points and repulsive forces to dissimilar ones. Subsequently, it gradually reduces the attractive and repulsive forces using simulated annealing optimization algorithm, as depicted in the following loss formulas:
\begin{equation}
A = {p}_{ij}(X)\log({\frac{{p}_{ij}(X)}{{q}_{ij}(Y)}})\label{eq6}
\end{equation}
\begin{equation}
   R = (1-{p}_{ij}(X))\log{\frac{1-{p}_{ij}(X)}{{1-{q}_{ij}(X)}}}\label{eq7}
\end{equation}
\begin{equation}
   CE(X,Y)=\displaystyle\sum_{i}\displaystyle\sum_{j}\left[A + R \right ]\label{eq8} 
\end{equation}
UMAP dimensionality reduction adjusts this formula to minimize the discrepancy between \({p}_{ij}\) and \({q}_{ij}\), aiming to achieve dimensional reduction.
\end{itemize}

\subsection{\textbf{PCA}}

Principal Component Analysis (PCA) is a widely used exploratory analysis method, particularly suitable for handling high-dimensional data \cite{b59}. Unlike many dimensionality reduction techniques, PCA has the advantage of not requiring any prior assumptions about the data, which means its implementation is relatively straightforward compared to methods like UMAP, making it particularly useful for processing new data. 

The PCA extracts the largest eigenvalues of the sample covariance matrix to ensure that new samples have maximum variance. 
\\Let \( {X}_{d\times n}=\begin{Bmatrix}
{x}_{1},...,{x}_{n}
\end{Bmatrix}\) represents \({x}_{ij}\), \( {Z}_{m\times n}=\begin{Bmatrix}
{z}_{1},...,{z}_{n}
\end{Bmatrix}\) represents the new samples. We need to find a metric \({L}_{m\times d}\) satisfying \(Z=LX\) having maximum variance.
\\Sample covariance matrix:
\begin{equation}
    {S}_{d\times d}=\frac{1}{n-1}X{X}^{T}\label{eq14}
\end{equation}
Use matrix decomposition to simplify formulate (15):
\begin{equation}
    {S}_{d\times d}=R\varLambda{R}^{-1}\label{eq15}
\end{equation}
where \(R=\begin{pmatrix}
{u}_{1},...,{u}_{d}
\end{pmatrix}\) is orthogonal matrix, \({u}_{i}\) represents the i-th eigenvector of \({S}\). \(\varLambda\) is diagonal matrix of the corresponding eigenvalues, \(\begin{vmatrix}
\varLambda
\end{vmatrix}=\displaystyle\prod_{i=1}^{d}{\lambda }_{i}\). 
\\PCA tends to choose the maximum m eigenvector corresponding to the largest m eigenvalues as principal component. Let \({L}_{d\times m}=\begin{pmatrix}
{u}_{1},...,{u}_{m}
\end{pmatrix}\), the transformation matrix \({L}_{m\times d}\) is obtained by taking the transpose of matrix \({L}_{d\times m}\).

\subsection{\textbf{t-SNE}}

t-Distributed Stochastic Neighbor Embedding (t-SNE) is a nonlinear dimensionality reduction algorithm that, like UMAP, maps high-dimensional data onto lower-dimensional spaces. This process seeks to reduce dimensions while preserving the local structure of the data. 

Assume \(X=\begin{Bmatrix}{x}_{1},...,{x}_{n}\mid {x}_{i}\epsilon {\mathbb{R}}^{k}\end{Bmatrix}\)  be a high dimensional input dataset and \(Y=\begin{Bmatrix}{y}_{1},...,{y}_{n}\mid {y}_{i}\epsilon {\mathbb{R}}^{M}\end{Bmatrix}\)  is the optimal low-dimensional data. The primary step of t-SNE is to transform the high-dimensional Euclidean distances between data points into conditional probabilities representing similarities. During this process, the similarity between data point \({x}_{i}\) and data point \({x}_{j}\) is expressed by the conditional probability \({p}_{j|i}\).
The conditional probability defined as: 
\begin{equation}
   {p}_{j|i}=\frac{e^{-\frac{\|{x}_{i}-{x}_{j}\|^{2}}{2{\sigma }_{i}^{2}}}}{\sum_{k\neq i}e^{-\frac{\|{x}_{i}-{x}_{k}\|^{2}}{2{\sigma }_{i}^{2}}}}\label{eq16}
\end{equation}

Taking into account the symmetry of distances between data points, the conditional probability is also symmetric. That is  \({p}_{j|i}={p}_{i|j}\). Define the probability as:
\begin{equation}
   {p}_{ij}= {p}_{j|i}+{p}_{i|j}\label{eq17} 
\end{equation}
 Through normalization, obtained:
 \begin{equation}
   {p}_{ij}= \frac{{p}_{ij}}{\displaystyle\sum_{i}\displaystyle\sum_{j}{p}_{ij}}\label{eq18} 
\end{equation}
 The corresponding joint probability of the data points mapped to the low-dimensional space is defined as:  
\begin{equation}
    {q}_{ij}=\frac{{\left ( 1+{\begin{Vmatrix}
{y}_{i}-{y}_{j}
\end{Vmatrix}}^{2}\right )}^{-1}}{\displaystyle\sum_{k\ne l}{\left ( 1+{\begin{Vmatrix}
{y}_{k}-{y}_{l}
\end{Vmatrix}}^{2}\right )}^{-1}}\label{eq19}
\end{equation}
Minimize the Kullback-Leibler (KL) divergence using gradient descent:
\begin{equation}
    KL\left ( P\parallel Q\right )= \displaystyle\sum_{i}\displaystyle\sum_{j}{p}_{ij}\log{\frac{{p}_{ij}}{{q}_{ij}}}\label{eq20}
\end{equation}

\subsection{\textbf{MLP}}
Multilayer Perceptrons (MLPs) \cite{b51} is a commonly used type of artificial neural network (ANN) designed for function approximation. It is capable of handling both linear and nonlinear functions, making it suitable for a variety of data relationships. Inspired by the functionality of the brain, the MLP is a powerful modeling tool. Through supervised training algorithms, MLPs can be trained with labeled data, allowing them to make accurate predictions even in previously unseen situations. 

The multilayer perceptron employs the Backpropagation (BP) algorithm for training. The BP algorithm is an optimization method used to train neural networks, aiming to minimize the loss function. By computing the discrepancy between the actual output and the desired output, the BP algorithm evaluates the model's performance. This discrepancy is represented by the loss function. The error signal is represented by the partial derivative of the loss function with respect to the weights of each layer. It is backpropagated through the network from the output layer to the input layer, utilizing the chain rule to compute this gradient. Then, the model parameters are updated using the Gradient Descent method, which involves subtracting the gradient multiplied by the learning rate from the current parameter values. This process is iteratively performed to minimize the error and optimise the model's performance. Ultimately, through backpropagation and gradient descent, the BP algorithm effectively adjusts the connection weights of the neural network, enabling the model to accurately map inputs to the desired outputs, optimizing its performance.

\subsection{\textbf{Method Discussion}}

The choice of dimensionality reduction techniques, including PCA, t-SNE, and UMAP, has long been a subject of debate due to their distinct advantages, limitations, and applicable scenarios.

PCA performs better than t-SNE and UMAP when reducing high-dimensional data to higher-dimensional subspaces. The reason is that PCA identifies directions of maximum variance in the data and projects it onto these directions to create a lower-dimensional space. By retaining as many principal components as possible, PCA can increase the explained variance ratio, which helps preserving the overall structure and variance information of the data. However, as a linear algorithm, PCA may not perform well when the data has nonlinear relationships.

When reducing data to lower-dimensional spaces, t-SNE and UMAP typically outperform PCA. t-SNE is a nonlinear method that works by finding a probability distribution of the high-dimensional data and then seeking a similar distribution in the lower-dimensional space, minimizing the difference between the two distributions. It effectively preserves the local structure of the data, clustering similar data points closely in the low-dimensional space. However, due to its high computational complexity and difficult-to-tune hyperparameters, its application is somewhat limited. UMAP, a newer nonlinear dimensionality reduction technique, is designed to preserve local structures while also maintaining global structures. Based on manifold learning theory, UMAP can process large datasets in a shorter time and performs well in preserving both local and global structures. Compared to t-SNE, UMAP runs faster and is less dependent on hyperparameters, making it more practical for many applications.

Becht et al. \cite{b4} in Nature Biotechnology argued that UMAP is superior to t-SNE because it better preserves the global structure of the data and is more consistent across runs. However, further research indicates that this perceived superiority is mainly due to different initialization methods. Becht et al. used random initialization for t-SNE by default, whereas UMAP was initialized using Laplacian eigenmaps. Studies show that UMAP with random initialization performs as poorly in preserving global structure as t-SNE with random initialization, whereas t-SNE with informative initialization performs as well as UMAP with informative initialization. Because of this reason, Becht et al.'s experiments do not demonstrate that the UMAP algorithm itself has an inherent advantage over t-SNE in preserving global structure \cite{b49}.

Consequently, the choice of dimensionality reduction method depends on the specific application scenario and data characteristics. Determining the optimal method based on data properties, computational resources, and specific analysis needs is a topic worth further investigation.

 \subsection{\textbf{ADRS-CNet}}
The adaptive classification function  of our model mainly relies on a specially trained multilayer perceptron classification model. The classification performance of a multilayer perceptron (MLP), as a fundamental neural network model, is significantly influenced by the appropriateness of the training and testing strategies employed. Therefore, devising scientifically sound training and testing strategies is crucial for enhancing the classification effectiveness of MLPs. By optimizing these strategies, the generalization ability and predictive accuracy of the model can be substantially improved.

To simplify the description, we define a process called "dataset sub-tagging". This process involves calculating the clustering accuracy of sub-datasets after dimensionality reduction using PCA, t-SNE, and UMAP. (Notice: To ensure comprehensive analysis, this study focuses on six target dimensions for dimensionality reduction, 2, 3, 50, 300, 500, and 700, which collectively span a wide range of the dimensional space.) Based on the obtained clustering accuracy, the dimensionality reduction method that yields the highest clustering accuracy for each sub-dataset is selected, and the corresponding sub-dataset is labelled with the tag of the most suitable dimensionality reduction method.

In the process of dataset sub-tagging, to more effectively compare the dimensionality reduction methods PCA, t-SNE, and UMAP, we fine-tuned the parameters for t-SNE. Specifically, we set the initial state of t-SNE to PCA, the maximum number of iterations to 300, the learning rate to 200, and adjusted the computation mode to “exact.” This adjustment aims to maximize the accuracy of t-SNE's dimensionality reduction while also enhancing its computational efficiency. The specific procedure is as follows:

We will divide the total dataset into an initial training set and an initial test set. Each initial training set will undergo random resampling, and these subsets will then be processed with dataset sub-tagging. This process will be repeated, and the subsets after sub-tagging will be aggregated into a unified group. (For each subset, we will sum the frequencies of each column in the k-mer frequency matrix of the DNA sequences, and use these frequency vectors, along with the target dimensions reduced using dimensionality reduction techniques during the dataset sub-tagging process, as feature vectors for classifying.) After balancing the labels and features using the SMOTE algorithm \cite{b52}, normalization and recursive feature elimination (RFE) \cite{b53}, these feature vectors and their labels will be used as a training set. We will then use the grid search method \cite{b54} to determine the optimal structure of the multilayer perceptron. Training these optimised MLP models, we can adaptively select the optimal dimensionality reduction method.

The testing process is similar to the training process. We randomly sampled 54 sub-test sets from the initial test set and input them into the trained MLP, with the objective of evaluating the model's weighted classification accuracy, recall, and F1-score. The optimal dimensionality reduction method identified by the MLP model was applied to the main model (a k-means clustering model based on the k-mer frequency matrix) to compute the average clustering accuracy of sub-test sets. By comparing the average clustering accuracy of the main model combined with PCA, UMAP, and t-SNE, we can more intuitively reflect the classification performance of our model. 

It is worth noting that our training and testing strategy draws inspiration from the Boosting classification algorithm \cite{b61}, where each model iteration builds on the information from previous iterations. We have integrated this concept into our MLP model, enabling it to learn from prior iterations. The specific steps are as follows:

We use the training set to train the model and then perform testing. Through identifying samples with unsatisfactory test results and reintroducing them into the training set for another round of training and testing, we iterate this process until the classification performance metrics reach the desired standards.   Theoretically, through such iterative testing and training, the classification performance will gradually improve with an increasing number of iterations. Since our method involves obtaining the test set through random sampling of the initial test set, the probability of having all identical DNA sequences in these subsets is nearly zero. This effectively reduces the risk of data leakage in our training and testing strategy. 

To facilitate comparison and evaluation, we have named our model "Adaptive Dimensionality Reduction Selection and Classification Network" (ADRS-CNet) based on its core functionality (Refer to Fig.2 for details).

\section{Experimental Results }

\subsection{\textbf{Dataset}}
The Clustered Nanopore Reads (CNR) Dataset \cite{b50} we used consists of two files:

\begin{itemize}
    \item \textbf{Centers.txt}: Contains 10,000 random DNA strings of length 110.
    \item \textbf{Clusters.txt}: Contains 269,709 noisy nanopore DNA reads corresponding to the strings in Centers.txt, organized into multiple clusters separated by lines of "====".
\end{itemize}

\subsection{\textbf{Evaluation Criteria}}

The evaluation metrics for the adaptive model classification are as follows  (To simplify the explanation, we use a binary classification problem as an example): 
\\For binary classification tasks, the outcomes of predictions and their corresponding true labels are classified into four categories: true positives (TP), true negatives (TN), false positives (FP), and false negatives (FN). 
\begin{itemize}

\item \textbf{Accuracy} 

\begin{equation}
\text{Accuracy} = \frac{\text{TP}}{\text{TP} + \text{FP}}\label{eq}
\end{equation}

\item \textbf{Recall}

\begin{equation}
\text{Recall} = \frac{\text{TP}}{\text{TP} + \text{FN}}\label{eq}
\end{equation}

\item \textbf{F1-Score}

\begin{equation}
\text{F1-Score} = 2 \times \frac{\text{Accuracy} \times \text{Recall}}{\text{Accuracy} + \text{Recall}}\label{eq}
\end{equation}
\end{itemize}
For the evaluation of a three-class classification task,three key averaging methods—macro average, micro average, and weighted average—are commonly used to summarize the model's performance. Considering the label distribution in the test set for dimensionality reduction methods in this experiment, we have decided to use weighted average precision, recall, and F1-score as the evaluation metrics for this study.
\begin{itemize}
\item \textbf{Clustering Accuracy} 
\end{itemize}
Since K-means clustering is an unsupervised learning algorithm, it does not have labels to explicitly evaluate clustering performance. However, we can test the accuracy of K-means clustering by using the labels from the dataset after dataset sub-tagging. For this purpose, we choose the number of clusters \( K \) to be equal to the number of categories in the sub-dataset and then compare the clustering results with the corresponding labels. This approach enables us to quantify the accuracy and, consequently, the performance of the dimensionality reduction-assisted K-means clustering method. 

For classification problems, we assume the training set is
\begin{equation}
D = \{(x_i, y_i)\}_{i=1}^n\label{eq} 
\end{equation}
where \( n \) is the number of samples, \( m \) is the number of features, and \( k \) is the number of labels. 

We set the number of clusters equal to the number of labels \( k \). After applying the K-means clustering algorithm, we obtain \( k \) different clusters \( \{c_1, c_2, \ldots, c_k\}\) . In each cluster \( c_j \), we define the K-means clustering predictor as:
\begin{equation}
     \text{predictor}(c_j) = \arg\max_{y_i \in \{y_1, \ldots, y_k\}} F_j(y_i)\label{eq}
\end{equation}
where \( F_j(y_i) \) is the frequency of label \( y_i \) in cluster \( c_j \).

The clustering accuracy can be defined as:
\begin{equation}
    \text{accuracy} = \frac{1}{n} \sum_{i=1}^n \mathbb{I}(\hat{y}_i = y_i)\label{eq}
\end{equation}
where \( \hat{y}_i \) is the predicted label, and \( \mathbb{I}(\hat{y}_i = y_i) \) is the indicator function, which equals 1 if \( \hat{y}_i = y_i \), otherwise 0.

\subsection{\textbf{Experimental Evaluation}}

To comprehensively validate the effectiveness of our model, our experiments are divided into three parts: testing on clusters 100 to 199, clusters 9800 to 9899, and randomly sampled clusters. These selected datasets are from "Clusters.txt" to form our total dataset, labeling each sample with its respective cluster label. Our classification model employs the optimal parameter combination obtained through grid search: two hidden layers with 50 neurons each, ReLU activation function, regularization coefficient of 0.05, learning rate of 0.001, and a maximum of 5000 iterations. This setup is designed to ensure the model's performance and generalization capability across different data distributions. 

We began our experiment by implementing the previously described training and testing strategy of ADRS-CNet, conducting three rounds of iterative testing. The classification performance of the MLP in each iteration was recorded in a table.  We then selected the iteration with the comparatively better classification performance to compare with the dimensionality reduction effects of PCA, t-SNE, and UMAP. The effectiveness of the chosen dimensionality reduction method was evaluated based on the clustering performance. 

\subsubsection{\textbf{Clusters 100 to 199}}

\begin{table}[H]
\centering
\captionsetup{font=small}
\caption{Classification Performance Metrics for ADRS-CNet }
\label{table:1}
\resizebox{0.5\textwidth}{!}{
\begin{tabular}{c c c c}
\toprule
\textbf{Cluster Range} & \multicolumn{3}{c}{100-199} \\
\midrule
\textbf{Iteration} & \textbf{First Iteration} & \textbf{Second Iteration} & \textbf{Third Iteration} \\
\midrule
\textbf{Weighted F1-score} & 0.748 & 0.865 & 0.895 \\
\textbf{Weighted Precision} & 0.780 & 0.861 & 0.970 \\
\textbf{Weighted Recall} & 0.772 & 0.870 & 0.852 \\
\textbf{Number of Important Features} & 3 & 4 & 4 \\
\bottomrule
\end{tabular}
}
\end{table}

\begin{table}[H]
\captionsetup{font=small}
    \centering
    \caption{First Ten Rows of Test Data and Classification Result}
    \label{tab:data}
    \renewcommand{\arraystretch}{1.2}
    \setlength{\tabcolsep}{1pt}
    \resizebox{0.5\textwidth}{!}{
        \begin{tabular}{lccccccccc}
            \toprule
            \textbf{n\_component} & \textbf{freq\_AAAAA} & \textbf{freq\_AAAAT} & \textbf{freq\_AAAAC} & \ldots & \textbf{best\_method} & \textbf{Predicted\_mlp} \\
            \midrule
            2   & 94  & 63  & 74    & \ldots & t-SNE  & t-SNE \\
            3   & 78  & 48  & 69    & \ldots & t-SNE  & t-SNE \\
            50  & 87  & 62  & 73   & \ldots & PCA  & PCA \\
            300 & 108 & 68  & 73  & \ldots & PCA  & PCA \\
            500 & 106 & 72  & 75   & \ldots & PCA  & UMAP \\
            700 & 102 & 64  & 77    & \ldots & PCA  & PCA \\
            2   & 98  & 72  & 60    & \ldots & t-SNE  & t-SNE \\
            3   & 83  & 61  & 62    & \ldots & t-SNE  & UMAP \\
            50  & 112 & 80  & 67   & \ldots & PCA  & PCA \\
            300 & 85  & 65  & 57  & \ldots & PCA  & PCA \\
            \bottomrule
        \end{tabular}
    }
\end{table}

\subsubsection{\textbf{Clusters 9800 to 9899}}

\begin{table}[H]
\captionsetup{font=small}
\centering
\caption{Classification Performance Metrics for ADRS-CNet }
\label{table:1}
\resizebox{0.5\textwidth}{!}{
\begin{tabular}{c c c c}
\toprule
\textbf{Cluster Range} & \multicolumn{3}{c}{9800-9899} \\
\midrule
\textbf{Iteration} & \textbf{First Iteration} & \textbf{Second Iteration} & \textbf{Third Iteration} \\
\midrule
\textbf{Weighted F1-score} & 0.613 & 0.854 & 0.945 \\
\textbf{Weighted Precision} & 0.729 & 0.942 & 0.928 \\
\textbf{Weighted Recall} & 0.648 & 0.796 & 0.963 \\
\textbf{Number of Important Features} & 2 & 3 & 3 \\
\bottomrule
\end{tabular}
}
\end{table}

\begin{table}[H]
\captionsetup{font=small}
    \centering
    \caption{First Ten Rows of Test Data and Classification Result}
    \label{tab:data}
     \renewcommand{\arraystretch}{1.2}
    \setlength{\tabcolsep}{1pt}
    \resizebox{0.5\textwidth}{!}{
    \begin{tabular}{lccccccccc}
        \toprule
        \textbf{n\_component} & \textbf{freq\_AAAAA} & \textbf{freq\_AAAAT} & \textbf{freq\_AAAAC}  & \ldots & \textbf{best\_method} & \textbf{Predicted\_mlp} \\
        \midrule
        2   & 144 & 93  & 90   & \ldots & t-SNE  & t-SNE \\
        3   & 139 & 89  & 87  & \ldots & t-SNE  & t-SNE \\
        50  & 136 & 95  & 92   & \ldots & PCA  & PCA \\
        300 & 146 & 100 & 87  & \ldots & PCA  & PCA \\
        500 & 118 & 101 & 75   & \ldots & PCA  & PCA \\
        700 & 124 & 88  & 80   & \ldots & PCA  & PCA \\
        2   & 134 & 96  & 93   & \ldots & t-SNE  & t-SNE \\
        3   & 142 & 90  & 95  & \ldots & t-SNE  & t-SNE \\
        50  & 144 & 98  & 80   & \ldots & PCA  & PCA \\
        300 & 142 & 98  & 80  & \ldots & PCA  & PCA \\
        \bottomrule
    \end{tabular}}
\end{table}

\subsubsection{\textbf{Random Clusters}}

\begin{table}[H]
\captionsetup{font=small}
\centering
\caption{Classification Performance Metrics for ADRS-CNet }
\label{table:1}
\resizebox{0.5\textwidth}{!}{
\begin{tabular}{c c c c}
\toprule
\textbf{Cluster Range} & \multicolumn{3}{c}{Random range} \\
\midrule
\textbf{Iteration} & \textbf{First Iteration} & \textbf{Second Iteration} & \textbf{Third Iteration} \\
\midrule
\textbf{Weighted F1-score} & 0.918 & 0.972 & 0.907 \\
\textbf{Weighted Precision} & 0.896 & 0.964 & 0.920 \\
\textbf{Weighted Recall} & 0.943 & 0.981 & 0.907 \\
\textbf{Number of Important Features} & 1 & 1 & 4 \\
\bottomrule
\end{tabular}
}
\end{table}

\begin{table}[H]
\captionsetup{font=small}
    \centering
    \caption{First Ten Rows of Test Data and Classification Result}
    \label{tab:data}
    \renewcommand{\arraystretch}{1.2}
    \setlength{\tabcolsep}{1pt}
    \resizebox{0.5\textwidth}{!}{
    \begin{tabular}{lccccccccc}
        \toprule
        \textbf{n\_component} & \textbf{freq\_AAAAA} & \textbf{freq\_AAAAT} & \textbf{freq\_AAAAC}  & \ldots & \textbf{best\_method} & \textbf{Predicted\_mlp} \\
        \midrule
        2   & 85  & 75 & 71  & \ldots & t-SNE  & t-SNE \\
        3   & 85  & 76  & 73  & \ldots & t-SNE  & t-SNE \\
        50  & 85  & 73 & 76  & \ldots & PCA  & PCA \\
        300 & 145  & 95  & 88 & \ldots & PCA & PCA \\
        500 & 132  & 85 & 86  & \ldots & PCA  & PCA \\
        700 & 122  & 85  & 92   & \ldots & PCA  & PCA \\
        2   & 74  & 89  & 92  & \ldots & UMAP  & UMAP \\
        3   & 83  & 85 & 97  & \ldots & UMAP  & t-SNE \\
        50  & 70  & 95  & 83   & \ldots & PCA  & PCA \\
        300 & 78 & 92  & 84  & \ldots & PCA  & PCA \\
        \bottomrule
    \end{tabular}}
\end{table}

Across the three experiments, although the classification performance may fluctuate due to the specific random sampling of the test sets, we consistently observed an improvement in performance as the number of iterations increased, provided that the feature selection was appropriate. Overall, our model maintained high weighted precision, recall, and F1-scores across the three iterations. High precision indicates that the model has fewer false positive errors, and high recall demonstrates that the model achieved high recall rates across the three classification categories. The F1-score, as a single metric balancing precision and recall, provides a comprehensive assessment of the model's accuracy. Based on the performance of these three classification metrics, we conclude that our classification model is highly effective. Particularly in the third random sampling experiment, the model effectively captured the patterns of the dimensionality reduction methods. This displays that our model performs exceptionally well in capturing the overall data structure (Refer to Table 1, 3,  and 5).

It should be emphasized that our model's classification errors in the low-dimensional space mainly occurred with the selection of t-SNE and UMAP, while in the high-dimensional space, higher error rates were mainly associated with PCA and UMAP selections. Nevertheless, in the low-dimensional space, the dimensionality reduction effects of t-SNE and UMAP are comparable, and in the high-dimensional space, the effects of PCA and UMAP are similar, both maintaining high levels of dimensionality reduction performance. This implies that even though our model may encounter some classification errors within these ranges, the average model accuracy remains at a high level, which is acceptable in practical applications (Refer to Table 2, 4 and 6).

We compared our model with three methods for dimensions reduction at six different dimensional scales (2, 3, 300, 500, and 700 dimensions) and calculated the average model accuracy for each method. These comparisons further validate the performance of our model across different dimensions (As shown in Fig.3, Fig.4, Fig.5, Fig.6, Fig.7 and Fig.8 ).

\section{Conclusion}
In this study, we introduce ADRS-CNet, an adaptive model for selecting dimensionality reduction methods within a K-means clustering framework.  This model is designed to optimise the dimensionality reduction process for DNA datasets, tackling the issues arising from the extensive dimensionality of k-mer frequency matrices. Our experiments with the Clustered Nanopore Reads (CNR) dataset show that ADRS-CNet not only effectively identifies and selects the optimal dimensionality reduction method but also achieves notable improvements in clustering accuracy compared to k-means clustering methods assisted by PCA, t-SNE, and UMAP. This adaptive model can effectively reduce noise and redundant information in DNA sequencing sequences, providing a potential solution for DNA storage sequence clustering analysis in the context of big data.

%%%%%%%%%%%%%%
\section{Appendices}
\begin{appendices}
\begin{figure}[H]
\centering
\includegraphics[width=0.4\textwidth,scale=0.5]{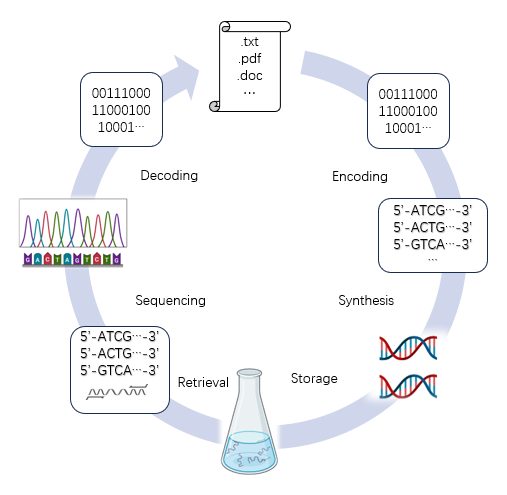}
    \caption{Major Processes of DNA Storage }
    \label{figure.1}
\end{figure}

\begin{figure}[H]
\centering
\includegraphics[width=0.5\textwidth,scale=0.5]{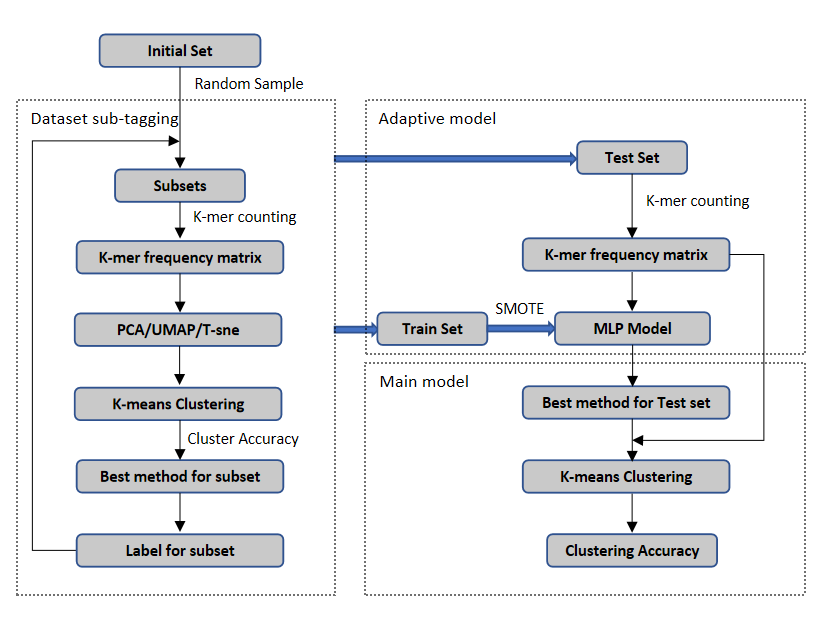}
    \caption{The framework for ADRS-CNet}
    \label{figure.1}
\end{figure}

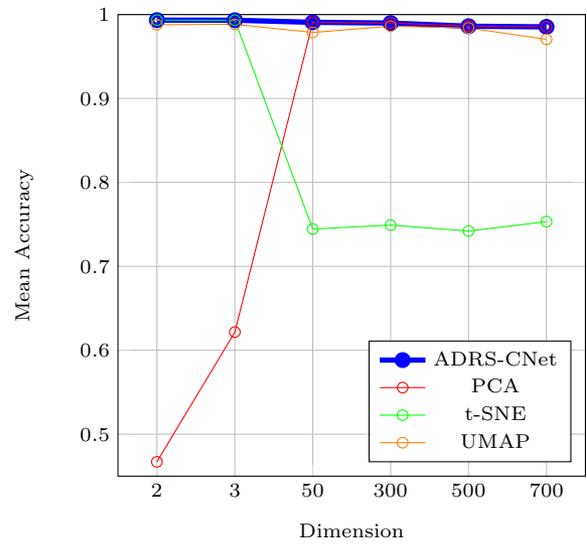
\begin{figure}[H]
    \centering
    \begin{tikzpicture}
        \begin{axis}[
            xlabel={Dimension},
            ylabel={ Mean Accuracy},
            grid=both,
            ymin=0.45, ymax=1.0,
            width=0.45\textwidth,
            height=0.45\textwidth,
            xtick={1,2,3,4,5,6},
            xticklabels={2,3,50,300,500,700},
            legend style={at={(0.97,0.03)}, anchor=south east}
        ]
        
        \addplot[
            color=blue,
            mark=o,
            line width=2pt
        ] coordinates {
            (1, 0.993381)
            (2, 0.993273)
            (3, 0.990777)
            (4, 0.989909)
            (5, 0.986111)
            (6, 0.985352)
        };
        \addlegendentry{ADRS-CNet }
        
        \addplot[
            color=red,
            mark=o
        ] coordinates {
            (1, 0.467014)
            (2, 0.621636)
            (3, 0.990777)
            (4, 0.989909)
            (5, 0.986220)
            (6, 0.985352)
        };
        \addlegendentry{PCA }
        
        \addplot[
            color=green,
            mark=o
        ] coordinates {
            (1, 0.993381)
            (2, 0.993273)
            (3, 0.744466)
            (4, 0.749132)
            (5, 0.742079)
            (6, 0.753255)
        };
        \addlegendentry{t-SNE}
        
        \addplot[
            color=orange,
            mark=o
        ] coordinates {
            (1, 0.987847)
            (2, 0.988390)
            (3, 0.978732)
            (4, 0.986111)
            (5, 0.984049)
            (6, 0.970486)
        };
        \addlegendentry{UMAP }
        
        \end{axis}
    \end{tikzpicture}
    \caption{100 to 199 Clustering accuracy with different dimensions}
    \label{fig:accuracy_dimension}
\end{figure}

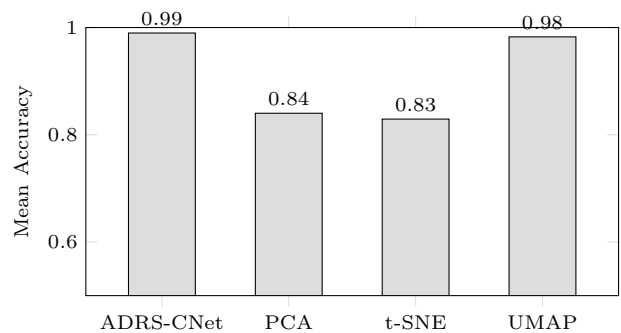
\begin{figure}[H]
    \centering
\begin{tikzpicture}
    \begin{axis}[
        ybar,
        ymin=0.5, ymax=1, % Set ymin to 0.5
        ylabel={ Mean Accuracy},
        symbolic x coords={ ADRS-CNet, PCA, t-SNE, UMAP},
        xtick=data,
        nodes near coords,
        bar width=25pt, % Increase bar width
        width=0.5\textwidth, % Set width to 0.5 times the width of the line
        height=0.30\textwidth, % Adjust height accordingly
        enlarge x limits={abs=1cm},
        ylabel near ticks,
        xlabel near ticks,
        nodes near coords style={color=black}, % Set the color of the annotations
    ]
        \addplot [fill=gray] coordinates {(ADRS-CNet,0.9898005) (PCA,0.840151) (t-SNE,0.829264) (UMAP,0.982603)};
    \end{axis}
\end{tikzpicture}
 \caption{100 to 199 Clustering accuracy with categorical axis}
    \label{fig:accuracy_dimension}
\end{figure}

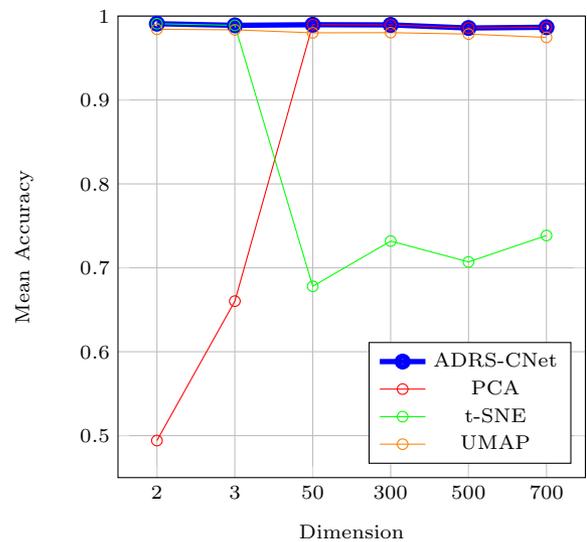
\begin{figure}[h!]
    \centering
    \begin{tikzpicture}
        \begin{axis}[
            xlabel={Dimension},
            ylabel={ Mean Accuracy},
            grid=both,
            ymin=0.45, ymax=1.0,
            width=0.45\textwidth,
            height=0.45\textwidth,
            xtick={1,2,3,4,5,6},
            xticklabels={2,3,50,300,500,700},
            legend style={at={(0.97,0.03)}, anchor=south east}
        ]
        \addplot[
            color=blue,
            mark=o,
            line width=2pt
        ] coordinates {
           
            (1, 0.990885)
            (2, 0.988932)
            (3, 0.989692)
            (4, 0.989475)
            (5, 0.985786)
            (6, 0.986654)    
        };
        \addlegendentry{ADRS-CNet }
        \addplot[
            color=red,
            mark=o
        ] coordinates {
(1, 0.494141)
            (2, 0.660265)
            (3, 0.989692)
            (4, 0.989475)
            (5, 0.985786)
            (6, 0.986654)
        };
        \addlegendentry{PCA }
        \addplot[
            color=green,
            mark=o
        ] coordinates {
           (1, 0.990885)
            (2, 0.988932)
            (3, 0.677951)
            (4, 0.731879)
            (5, 0.707031)
            (6, 0.738498)
        };
        \addlegendentry{t-SNE }
        \addplot[
            color=orange,
            mark=o
        ] coordinates {
           (1, 0.984266)
            (2, 0.983615)
            (3, 0.980035)
            (4, 0.980252)
            (5, 0.978516)
            (6, 0.974609)
        };
        \addlegendentry{UMAP }
        \end{axis}
    \end{tikzpicture}
    \caption{9800 to 9899 clustering accuracy with different dimensions}
    \label{fig:accuracy_dimension}
\end{figure}

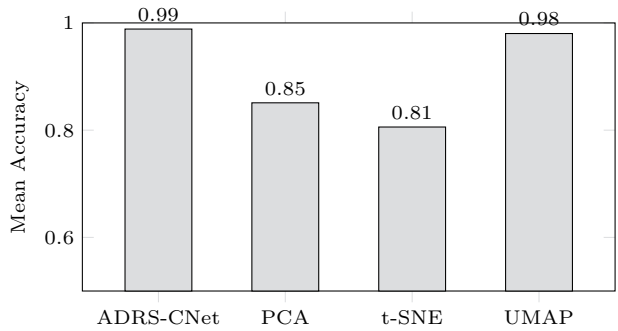
\begin{figure}[h!]
    \centering
\begin{tikzpicture}
    \begin{axis}[
        ybar,
        ymin=0.5, ymax=1, % Set ymin to 0.5
        ylabel={ Mean Accuracy},
        symbolic x coords={ADRS-CNet, PCA, t-SNE, UMAP},
        xtick=data,
        nodes near coords,
        bar width=25pt, % Increase bar width
        width=0.5\textwidth, % Set width to 0.5 times the width of the line
        height=0.30\textwidth, % Adjust height accordingly
        enlarge x limits={abs=1cm},
        ylabel near ticks,
        xlabel near ticks,
        nodes near coords style={color=black}, % Set the color of the annotations
    ]
        \addplot [fill=gray] coordinates{(ADRS-CNet,0.988571) (PCA,0.851002) (t-SNE,0.805863) (UMAP,0.980216)};
    \end{axis}
\end{tikzpicture}
 \caption{9800 to 9899 clustering accuracy with categorical axis}
    \label{fig:accuracy_dimension}
\end{figure}

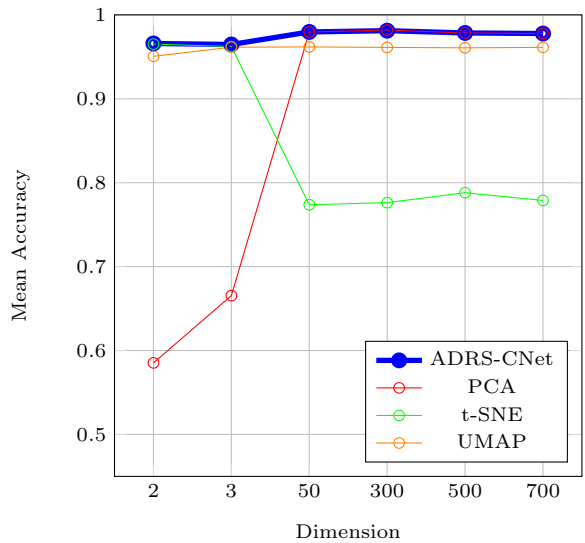
\begin{figure}[h!]
    \centering
    \begin{tikzpicture}
        \begin{axis}[
            xlabel={Dimension},
            ylabel={ Mean Accuracy},
            grid=both,
            ymin=0.45, ymax=1.0,
            width=0.45\textwidth,
            height=0.45\textwidth,
            xtick={1,2,3,4,5,6},
            xticklabels={2,3,50,300,500,700},
            legend style={at={(0.97,0.03)}, anchor=south east}
        ]
        \addplot[
            color=blue,
            mark=o,
            line width=2pt
        ] coordinates {
           (1, 0.965895) (2, 0.964693) (3, 0.979642) (4, 0.981145) (5, 0.978440) (6, 0.977764)
        };
        \addlegendentry{ADRS-CNet }
        \addplot[
            color=red,
            mark=o
        ] coordinates {
            (1, 0.585261) (2, 0.665415) (3, 0.979642) (4, 0.981896) (5, 0.978440) (6, 0.977764)
        };
        \addlegendentry{PCA}
        \addplot[
            color=green,
            mark=o
        ] coordinates {
          (1, 0.965144) (2, 0.962215) (3, 0.773738) (4, 0.776217) (5, 0.788086) (6, 0.778846)
        };
        \addlegendentry{t-SNE}
        
        \addplot[
            color=orange,
            mark=o
        ] coordinates {
            (1, 0.950571) (2, 0.961463) (3, 0.961764) (4, 0.961238) (5, 0.960712) (6, 0.961238)
        };
        \addlegendentry{UMAP}
        
        \end{axis}
    \end{tikzpicture}
    \caption{Random clustering accuracy with different dimensions}
    \label{fig:accuracy_dimension}
\end{figure}

\begin{figure}[h!]
    \centering
\begin{tikzpicture}
    \begin{axis}[
        ybar,
        ymin=0.5, ymax=1, % Set ymin to 0.5
        ylabel={ Mean Accuracy},
        symbolic x coords={ADRS-CNet, PCA, t-SNE, UMAP},
        xtick=data,
        nodes near coords,
        bar width=25pt, % Increase bar width
        width=0.5\textwidth, % Set width to 0.5 times the width of the line
        height=0.30\textwidth, % Adjust height accordingly
        enlarge x limits={abs=1cm},
        ylabel near ticks,
        xlabel near ticks,
        nodes near coords style={color=black}, % Set the color of the annotations
    ]
        \addplot [fill=gray] coordinates {(ADRS-CNet, 0.9745965) (PCA, 0.861403) (t-SNE, 0.84070767) (UMAP, 0.95949767)};
    \end{axis}
\end{tikzpicture}
 \caption{Random clustering accuracy with categorical axis}
    \label{fig:accuracy_dimension}
\end{figure}
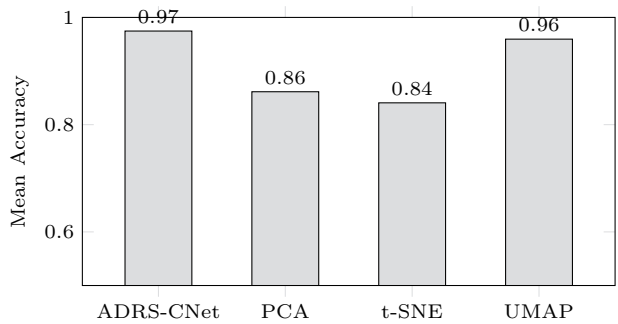

\end{appendices}

\bibliographystyle{plain}
\bibliography{reference}

\begin{thebibliography}{10}

\bibitem{b17}
A.~Alharthi, J.~Krotov, and N.~Bowman.
\newblock Addressing barriers to big data.
\newblock {\em Business Horizons}, 60(3):285--292, 2017.

\bibitem{b45}
Z.~Bao, J.~Zhu, H.~Yan, et~al.
\newblock An improved alignment-free model for dna sequence similarity metric.
\newblock {\em BMC Bioinformatics}, 15(1):1--12, 2014.

\bibitem{b4}
E.~Becht, L.~McInnes, J.~Healy, et~al.
\newblock Dimensionality reduction for visualizing single-cell data using umap.
\newblock {\em Nature Biotechnology}, 37(1):38--44, 2018.

\bibitem{b51}
C.~M. Bishop.
\newblock {\em Neural networks for pattern recognition}.
\newblock Oxford University Press, Oxford, U.K., 1995.

\bibitem{b12}
L.~Ceze, J.~Nivala, and K.~Strauss.
\newblock Molecular digital data storage using dna.
\newblock {\em Nature Reviews Genetics}, 20:456--466, 2019.

\bibitem{b34}
S.~Chandak, P.~Tatwawadi, J.~Koch, et~al.
\newblock Improved read/write cost tradeoff in dna-based data storage using ldpc codes, 2019.
\newblock bioRxiv. doi:10.1101/770032.

\bibitem{b41}
T.~Chappell, S.~Geva, and J.~Hogan.
\newblock K-means clustering of biological sequences.
\newblock In {\em Proceedings of the 22nd Australasian Document Computing Symposium}, 2017.

\bibitem{b52}
N.~V. Chawla, K.~W. Bowyer, L.~O. Hall, and W.~P. Kegelmeyer.
\newblock Smote: Synthetic minority over-sampling technique.
\newblock {\em Journal of Artificial Intelligence Research}, 16:321--357, 2002.
\newblock Available at: \url{https://www.jair.org/index.php/jair/article/view/10302}.

\bibitem{b53}
X.-w. Chen and J.~C. Jeong.
\newblock Enhanced recursive feature elimination.
\newblock In {\em Proceedings of the 6th International Conference on Machine Learning Applications (ICMLA 2007)}, pages 429--435, Cincinnati, OH, USA, 2007.

\bibitem{b16}
C.~de~Boer and A.~Regev.
\newblock Brockman: Deciphering variance in epigenomic regulators by k-mer factorization.
\newblock {\em BMC Bioinformatics}, 19(1), 2018.

\bibitem{b9}
Y.~Dong, F.~Sun, Z.~Ping, et~al.
\newblock Dna storage: research landscape and future prospects.
\newblock {\em National Science Review}, 7(6):1092--1107, 2020.

\bibitem{b38}
R.~C. Edgar.
\newblock Search and clustering orders of magnitude faster than blast.
\newblock {\em Bioinformatics}, 26(19):2460–2461, 2010.

\bibitem{b40}
R.~C. Edgar.
\newblock Uclust algorithm, 2024.
\newblock Available at: \url{https://www.drive5.com/usearch/manual/uclust_algo.html} [Accessed 10 July 2024].

\bibitem{b33}
Y.~Erlich and D.~Zielinski.
\newblock Dna fountain enables a robust and efficient storage architecture.
\newblock {\em Science}, 355(6328):950--954, 2017.

\bibitem{b13}
A.~Extance.
\newblock How dna could store all the world's data.
\newblock {\em Nature}, 537:22--24, 2016.

\bibitem{b54}
M.~Feurer and F.~Hutter.
\newblock Hyperparameter optimization.
\newblock In F.~Hutter, L.~Kotthoff, and J.~Vanschoren, editors, {\em Automated Machine Learning}, pages 3--33. Springer, Cham, Switzerland, 2019.

\bibitem{b61}
Yoav Freund and Robert~E. Schapire.
\newblock Experiments with a new boosting algorithm.
\newblock In {\em ICML'96: Proceedings of the Thirteenth International Conference on International Conference on Machine Learning}, pages 148--156, San Francisco, CA, United States, 1996. Morgan Kaufmann Publishers Inc.

\bibitem{b43}
O.~Gotoh.
\newblock An improved algorithm for matching biological sequences.
\newblock {\em Journal of Molecular Biology}, 162(3):705–708, 1982.

\bibitem{b8}
Y.~Hozumi, R.~Wang, C.~Yin, and G.-W. Wei.
\newblock Umap-assisted k-means clustering of large-scale sars-cov-2 mutation datasets.
\newblock {\em Computers in Biology and Medicine}, 131:104264, 2021.

\bibitem{b47}
Y.~Hrytsenko, N.~M. Daniels, and R.~S. Schwartz.
\newblock Determining population structure from k-mer frequencies, 2022.
\newblock Research Square, preprint. doi:10.21203/rs.3.rs-1689838/v2.

\bibitem{b46}
B.~Johnson, C.~Salinas, C.~Nelson, et~al.
\newblock Meshclust: an intelligent tool for clustering dna sequences.
\newblock {\em Nucleic Acids Research}, 46(14):e83--e83, 2018.

\bibitem{b59}
I.~T. Jolliffe and J.~Cadima.
\newblock Principal component analysis: a review and recent developments.
\newblock {\em Philosophical Transactions of the Royal Society A: Mathematical, Physical and Engineering Sciences}, 374(2065):20150202, 2016.

\bibitem{b49}
D.~Kobak and G.~C. Linderman.
\newblock Initialization is critical for preserving global data structure in both t-sne and umap.
\newblock {\em Nature Biotechnology}, 39(2):156--157, 2021.

\bibitem{b35}
W.~Li and A.~Godzik.
\newblock Cd-hit: a fast program for clustering and comparing large sets of protein or nucleotide sequences.
\newblock {\em Bioinformatics}, 22(13):1658--1659, 2006.

\bibitem{b7}
A.~Maćkiewicz and W.~Ratajczak.
\newblock Principal components analysis (pca).
\newblock {\em Computers \& Geosciences}, 19(3):303--342, 1993.

\bibitem{b5}
L.~McInnes, J.~Healy, and J.~Melville.
\newblock Umap: Uniform manifold approximation and projection for dimensionality reduction, 2024.
\newblock Available at: \url{https://arxiv.org/abs/1802.03426} [Accessed 6 July 2024].

\bibitem{b20}
Y.~Nakamura, K.~Oshima, T.~Moriya, et~al.
\newblock Sequence-specific error profile of illumina sequencers.
\newblock {\em Nucleic Acids Research}, 39(13):e90--e90, 2011.

\bibitem{b37}
S.~B. Needleman and C.~D. Wunsch.
\newblock A general method applicable to the search for similarities in the amino acid sequence of two proteins.
\newblock {\em Journal of Molecular Biology}, 48(3):443--453, 1970.

\bibitem{b19}
Ş. Ozan.
\newblock Dna sequence classification with compressors, 2024.
\newblock Available at: \url{https://arxiv.org/html/2401.14025v1} [Accessed 6 July 2024].

\bibitem{b10}
Z.~Ping, D.~Ma, X.~Huang, et~al.
\newblock Carbon-based archiving: current progress and future prospects of dna-based data storage.
\newblock {\em GigaScience}, 8:giz076, 2019.

\bibitem{b44}
J.~A.~M. Rexie, K.~Raimond, D.~Brindha, and A.~K. Prabavathy.
\newblock K-mer based prediction of gene family by applying multinomial naïve bayes algorithm in dna sequence.
\newblock In {\em AIP Conference Proceedings}, volume 2914, page 050025, 2023.

\bibitem{b32}
S.~Sankar, H.~Sun, J.~Gehring, A.~Singhal, and C.~M. Ajo-Franklin.
\newblock Comparative analysis of clustering methodologies in dna storage.
\newblock In {\em Proceedings of the 2022 26th International Computer Science and Engineering Conference (ICSEC)}, 2022.

\bibitem{b11}
Seagate.
\newblock Data age 2025, 2024.
\newblock Available at: \url{https://www.seagate.com/files/www-content/our-story/trends/files/idc-seagate-dataage-whitepaper.pdf} [Accessed 6 July 2024].

\bibitem{b15}
N.~Singh.
\newblock Demystify dna sequencing with machine learning, 2024.
\newblock Available at: \url{https://www.kaggle.com/code/nageshsingh/demystify-dna-sequencing-with-machine-learning/notebook} [Accessed 6 July 2024].

\bibitem{b50}
S.~R. Srinivasavaradhan, S.~Gopi, H.~D. Pfister, and S.~Yekhanin.
\newblock Trellis bma: Coded trace reconstruction on ids channels for dna storage.
\newblock In {\em Proceedings of the 2021 IEEE International Symposium on Information Theory (ISIT)}, pages 2453--2458, Melbourne, Australia, 2021.

\bibitem{b22}
L.~Van~der Maaten and G.~Hinton.
\newblock Visualizing data using t-sne.
\newblock {\em Journal of Machine Learning Research}, 9(86):2579--2605, 2008.

\bibitem{b24}
S.~Wilkinson.
\newblock Introduction to the kmer r package, 2024.
\newblock Available at: \url{https://cran.r-project.org/web/packages/kmer/vignettes/kmer-vignette.html} [Accessed 7 July 2024].

\bibitem{b3}
P.~Xu.
\newblock Dna storage and its research progress, 2024.
\newblock Available at: \url{https://www.researchgate.net/publication/342945230_DNA_Storage_and_Its_Research_Progress} [Accessed 6 July 2024].

\bibitem{b14}
V.~Zhirnov, R.~M. Zadegan, G.~S. Sandhu, et~al.
\newblock Nucleic acid memory.
\newblock {\em Nature Materials}, 15:366--370, 2016.

\bibitem{b6}
Y.-Y. Zhu.
\newblock Dna sequence data mining technique.
\newblock {\em Journal of Software}, 18(11):2766, 2007.

\bibitem{b36}
E.~Zorita, P.~Cusco, and G.~J. Filion.
\newblock Starcode: sequence clustering based on all-pairs search.
\newblock {\em Bioinformatics}, 31(12):1913--1919, 2015.

\end{thebibliography}

%USE THE BELOW OPTIONS IN CASE YOU NEED AUTHOR YEAR FORMAT.
%\bibliographystyle{abbrvnat}
%\bibliography{reference}

%% sample for biography with author's image
%\begin{biography}{{\color{black!20}\rule{77pt}{77pt}}}{\author{Author Name.} This is sample author biography text. The values provided in the optional argument are meant for sample purposes. There is no need to include the width and height of an image in the optional argument for live articles. This is sample author biography text this is sample author biography text this is sample author biography text this is sample author biography text this is sample author biography text this is sample author biography text this is sample author biography text this is sample author biography text.}
%\end{biography}

%% sample for biography without author's image
%\begin{biography}{}{\author{Jiankun Li} This is sample author biography text this is sample author biography text this is sample author biography text this is sample author biography text this is sample author biography text this is sample author biography text this is sample author biography text this is sample author biography text.}
%\end{biography}

\end{document}